\documentclass[journal]{IEEEtran}
\usepackage{url}
\usepackage{amssymb,amsmath,bm}
\usepackage[caption=false,font=normalsize,labelfont=sf,textfont=sf]{subfig}
\usepackage{hyperref,color}
\hypersetup{
    colorlinks=true,
    linkcolor=blue,
    urlcolor=blue,
    citecolor=cyan,
}
\usepackage{balance}
\usepackage{multirow}
\usepackage{graphicx}

\ifCLASSINFOpdf
\else
\fi
\begin{document}
%
\title{Uncertain Facial Expression Recognition via Multi-task Assisted Correction}
%
%
%

\author{Yang~Liu, 
        Xingming~Zhang, 
        Janne Kauttonen, 
        and~Guoying~Zhao$^*$,~\IEEEmembership{Fellow,~IEEE}
\thanks{$^*$ Corresponding author}
\thanks{Y. Liu and G. Zhao are with the Center for Machine Vision and Signal Analysis, University of Oulu, Finland, FI-90014 e-mail: Yang.Liu@oulu.fi, Guoying.Zhao@oulu.fi}
\thanks{X. Zhang is with the School of Computer Science and Engineering, South China University of Technology, Guangzhou, China, 510006 e-mail: cxzxm@scut.edu.cn}
\thanks{J. Kauttonen is with the Haaga-Helia University of Applied Sciences, Helsinki, Finland, FI-00520 e-mail: Janne.Kauttonen@haaga-helia.fi}}

%
%

\markboth{Journal of \LaTeX\ Class Files,~Vol.~14, No.~8, August~2015}%
{Shell \MakeLowercase{\textit{et al.}}: Bare Demo of IEEEtran.cls for IEEE Journals}
%



\maketitle

\begin{abstract}
Deep models for facial expression recognition achieve high performance by training on large-scale labeled data. However, publicly available datasets contain uncertain facial expressions caused by ambiguous annotations or confusing emotions, which could severely decline the robustness. Previous studies usually follow the bias elimination method in general tasks without considering the uncertainty problem from the perspective of different corresponding sources. In this paper, we propose a novel method of multi-task assisted correction in addressing uncertain facial expression recognition called MTAC. Specifically, a confidence estimation block and a weighted regularization module are applied to highlight solid samples and suppress uncertain samples in every batch. In addition, two auxiliary tasks, i.e., action unit detection and valence-arousal measurement, are introduced to learn semantic distributions from a data-driven AU graph and mitigate category imbalance based on latent dependencies between discrete and continuous emotions, respectively. Moreover, a re-labeling strategy guided by feature-level similarity constraint further generates new labels for identified uncertain samples to promote model learning. The proposed method can flexibly combine with existing frameworks in a fully-supervised or weakly-supervised manner. Experiments on RAF-DB, AffectNet, and AffWild2 datasets demonstrate that the MTAC obtains substantial improvements over baselines when facing synthetic and real uncertainties and outperforms the state-of-the-art methods. 
\end{abstract}


%
\IEEEpeerreviewmaketitle

\section{Introduction}
\IEEEPARstart{F}{acial} expressions carry essential information for perceiving human emotions and attitudes in daily communications. Automatic facial expression recognition (FER) from visual signals of images and videos is a vital technology for realizing human-computer systems such as remote health care, virtual reality, and social robots. Due to sufficient labeled data and high-speed computation resources, deep learning models have achieved excellent performance and dominated the FER research in recent years \cite{zhang2022median, li2022deep, li2022learning}. 

High-quality annotated images are significant when developing a FER method. Early facial expression datasets (e.g., CK+ \cite{lucey2010extended} and Oulu-CASIA \cite{zhao2011facial}) usually recruit a small scale of subjects and collect their facial expressions in a lab-controlled environment. Due to the limited number and conditions, experts can annotate the data carefully and precisely. To meet the requirement of massive labeled samples for training a deep FER model, recently released datasets (e.g., RAF-DB \cite{li2019reliable}, AffectNet \cite{mollahosseini2017affectnet}, and AffWild2 \cite{kollias2022abaw}) gather images from the Internet. For those real-world datasets, their annotations are difficult to maintain consistency in a large-scale manner. As a result, many labels are uncertain or even incorrect, which may cause two negative impacts on model learning. First, the over-fitting problem will arise due to the considerable proportion of ambiguous samples in the training set. Second, the uncertain labels will mislead the model learning features of specific facial expressions and decrease recognition performance.

\begin{figure}[t]
   \centering
   \includegraphics[width=0.85\columnwidth]{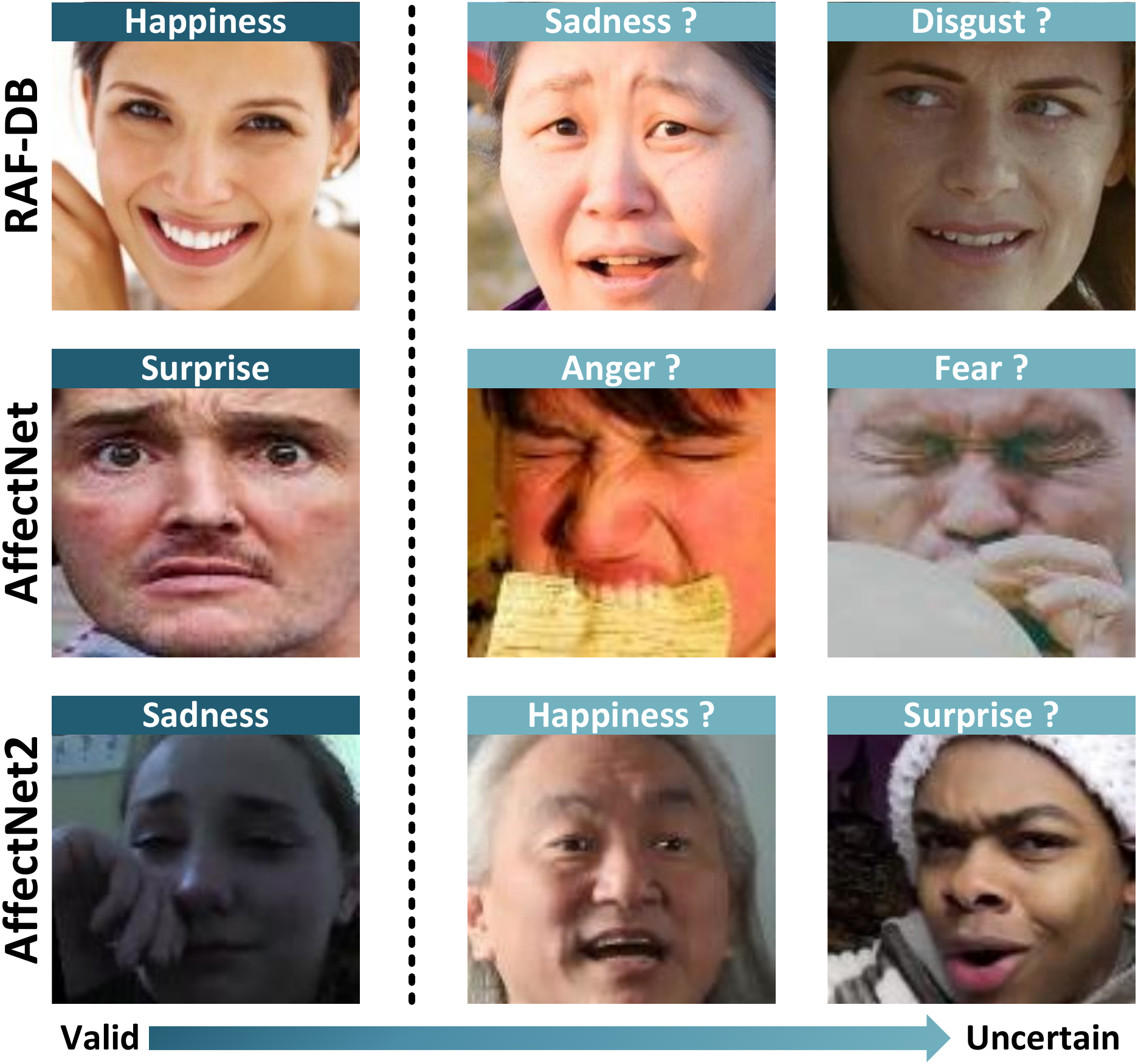}
   \caption{Examples of valid and uncertain images in RAF-DB, AffectNet, and AffWild2 datasets. Top texts indicate their original labels. Uncertainty commonly exists in different facial expression datasets.}\label{fig:uncertainty}
\end{figure}

The uncertainty can be divided into two categories in the FER task according to the source. The first one is subjective annotation. Labels for existing facial expression datasets are voted by annotators recruited on crowdsourcing platforms \cite{li2019reliable}. These annotators usually do not have the expertise and will assign different labels for the same image based on their backgrounds, especially for facial expressions under in-the-wild scenes. In Fig. \ref{fig:uncertainty}, we show several examples in RAF-DB and AffectNet datasets to illustrate the prevalence of uncertain samples. For samples on the left column, annotators can easily make consistent labeling. While for the right column, it is evident that multiple annotators might have various perspectives on the same sample. 

\begin{figure}[t]
   \centering
   \includegraphics[width=1\columnwidth]{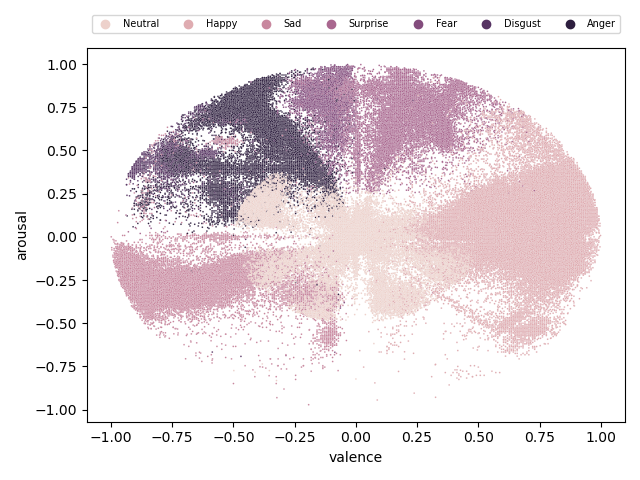}
   \caption{Distribution of discrete labels and continuous labels in AffectNet's training set. Categories with few samples are more likely to be confused with other classes. Many samples are far away from the center of the original category but close to other clusters, indicating a great deal of uncertainty.}\label{fig:scatter}
\end{figure}

The second source is intrinsic confusion. Existing FER methods usually predict basic emotions (e.g., anger, disgust, fear, happiness, neutral, sadness, and surprise). However, expression behaviors in daily life are spontaneous and diverse due to different induction, posture, and context \cite{liu2018visual, kollias2019deep}. Many facial expressions consist of compound or non-basic emotions under in-the-wild scenarios, which are difficult to be described by discrete labels. This phenomenon becomes even worse when encountering the class imbalance problem in the dataset, as shown in the distribution visualization of training labels in the AffectNet dataset (see Fig. \ref{fig:scatter}). 

To this end, a few studies have proposed solutions to alleviate the uncertainty problem. Most of them focused on migrating the methods of handling data noise in general tasks \cite{she2021dive,li2021learning}. Generally, a specific block for uncertainty estimation will be introduced to weight or relabel every sample during the model training \cite{gu2022towards}. Recently, considering characteristics of the FER task in terms of the variety of annotations and the inter-connectivity of sub- or similar tasks, the idea of using the relationship among multiple labels such as action units (AUs) and valence-arousal (VA) has been explored \cite{cui2020label,chen2020label}. However, these methods are still plagued by uncertain samples for the following reasons: 1) different types of uncertainty are lumped together without targeted treatment; 2) although additional multi-label knowledge is applied, semantic information is considered from the label level rather than the feature level; 3) the relabeling strategy without constraints is usually coarse, which could decrease the reliability of the generated labels.

In this paper, we develop a new framework to perform uncertain FER via \textbf{M}ulti-\textbf{T}ask \textbf{A}ssisted \textbf{C}orrection, called MTAC. It consists of three parts: a target branch and two auxiliary branches. Facial features are first extracted through a backbone network for every batch of training data. A weighted regularization module estimates each sample by learning confidence and encourages the model to focus on images with reliable labels in the target branch. Based on a parameter-sharing backbone, the VA estimation task is introduced to jointly supervise the feature learning with a consideration of category imbalance, while the AU detection task is conducted by adding a graph convolution block and extracting the semantic representation of each sample. For those samples that are identified as extremely uncertain, we compare their semantic representation with memory templates and relabel them under the constraint of feature-level similarity.

This paper is an extended work of our preliminary study published on ICPR 2022 \cite{liu2022uncertain}. Compared to the previous version, we have enhanced this paper in four aspects: 1) we involve a new auxiliary task of VA estimation for collaborative model training, which can address the uncertainty caused by the biases of discrete labels on describing in-the-wild facial expressions; 2) we revise original loss functions and design a new weighted loss for handling data imbalance, which can jointly optimize the feature extractor; 3) we construct a new memory template of weighted semantic centers and improve the relabeling strategy, which can adaptively generate pseudo labels for uncertain samples; 4) we employ additional backbones and datasets for more comprehensive experiments to evaluate the effectiveness of our method. Overall, the main contributions of this paper are summarized as follows:

\begin{itemize}
  \item MTAC method quantifies the sample confidence and suppresses the effects of uncertain discrete labels during the model training.
  \item MTAC mitigates the category imbalance and facilitates feature learning on ambiguous facial expressions with continuous labels in the auxiliary VA estimation task.
  \item MTAC explores the semantic representation from the auxiliary AU detection task and conducts uncertain label correction under a feature-level constraint.
  \item Extensive experiments on three large-scale datasets demonstrate that our MTAC can effectively solve the uncertain problem and achieves superior performance against state-of-the-art methods.
\end{itemize}

The rest of this paper is organized as follows: Section \ref{sec:rw} reviews several related studies, Section \ref{sec:pm} elaborates on the proposed MTAC method, Section \ref{sec:exp} reports the experimental results and discussions, and Section \ref{sec:end} concludes this work.

\section{Related Work}\label{sec:rw}
In this section, we briefly summarize the recent progress of the FER research in terms of multi-task facial expression analysis, graph-based affective representation, and deep learning with uncertainty.

\subsection{Multi-task facial expression analysis}
Automatically predicting basic emotions is the main task in traditional FER studies. Benefiting from psychological findings, advanced emotional description models have been utilized to annotate a broader range of facial expressions, such as AUs \cite{friesen1978facial} and VA \cite{russell1978evidence}. Therefore, recent studies have explored combining multiple tasks for a generalized feature extractor of facial expressions. Chen and Joo \cite{chen2021understanding} incorporated the \emph{triplet} loss into the objective function to embed the dependency between AUs and expression categories. Zhang \emph{et al.} \cite{zhang2021facial} designed a unified adversarial learning framework to link emotion prediction and the joint distribution of dimensional labels. Similarly, Antoniadis \emph{et al.} \cite{antoniadis2021exploiting} captured the dependencies between categorical and dimensional emotions through a graph convolutional network (GCN). Cui \emph{et al.} \cite{cui2020label} extracted the dependency between object-level labels and property-level labels, which could be used to revise and generate labels for new datasets. Besides emotion-related tasks, other close facial tasks like facial landmark detection have been proven to provide additional information to facial expression analysis. Chen \emph{et al.} \cite{chen2020label} introduced landmark detection as a neighbor task and leveraged the distribution of cluster samples to handle the label inconsistency. Toisoul \emph{et al.} \cite{toisoul2021estimation} integrated facial landmarks with discrete and continuous emotions into a single network that features around fiducial points were used to build an attention mechanism. Unlike previous methods, we exploit AU detection and VA estimation as two auxiliary tasks to assist uncertainty correction in this paper. Each auxiliary task can be independently integrated during the model training without causing extra burden in the testing stage. 

\subsection{Graph-based affective representation}
Effective facial representations are vital for FER methods. Recently, graph-based methods have been proposed because they can simultaneously represent facial anatomy and semantic relationships among facial areas, which are considered crucial clues of human facial perception \cite{liu2022graph}. Liu \emph{et al.} \cite{liu2021sg} designed a graph representation of facial expressions that consisted of reasonable facial landmarks and semantic connections, which modeled critical appearance and geometric facial changes. Zhao \emph{et al.} \cite{zhao2021geometry} constructed a geometric graph description of facial components that were more robust to appearance variations like texture noise and light changes. Besides facial landmarks, many studies generate graph representations based on local facial regions. Jin \emph{et al.} \cite{jin2021learning} cropped 20 local facial areas as graph nodes and linked edges according to a trainable weighted adjacency matrix to exploit intra- and inter-regional relationships. Xie \emph{et al.} \cite{xie2020adversarial} correlated a cross-domain graph for global-local feature adaptation to learn invariant representations of facial expressions. Alternatively, graphs constructed from the perspective of AUs are also explored. Luo \emph{et al.} \cite{luo2022learning} learned a unique graph that described the relationship between each pair of AUs, including its activation status and its association with other AUs. Song \emph{et al.} \cite{song2021hybrid} transferred hybrid messages among AUs and inferred possible graph structures to provide complementary information for higher performance. In this work, we focus on AU graphs where the extracted semantic representation is used to constrain the re-labeling strategy. Compared to existing methods, our AU graph is built based on a data-driven way in a fully supervised or a weakly-supervised manner.

\subsection{Deep learning with label uncertainty}
Label uncertainty is a common and significant problem in FER, and plagues deep models for many general tasks \cite{song2022learning}. Machine learning researchers usually regard uncertainty as a noisy label issue and rely on modified loss functions to penalize it. Zhong \emph{et al.} \cite{zhong2019graph} propagated the uncertain signals across a confidence graph based on feature similarity and temporal consistency that were used to train a label noise cleaner. Li \emph{et al.} \cite{li2021learning} regularized the low-dimensional subspace of embedded images by a consistency loss and a prototypical loss so that alleviated uncertain samples with a neighboring constraint. Analogously, in the FER field, Wang \emph{et al.} \cite{wang2020suppressing} proposed a self-cure network to learn the importance weight of each facial image and suppress uncertain samples by identifying and modifying untruthful labels. She \emph{et al.} \cite{she2021dive} exploited auxiliary multi-branch distribution learning and pairwise uncertainty estimation to solve the ambiguity in both the label space and the instance space. Zhang \emph{et al.} \cite{zhang2021weakly} formulated a noise modeling network based on a weakly-supervised strategy that learned the mapping from feature space to the residuals between clean and noisy labels. Gu \emph{et al.} \cite{gu2022towards} suppressed the label and feature noise by leveraging a multivariate normal distribution and preserving the inter-class correlations. As mentioned before, the FER task suffers from various uncertainties, which need to be fully considered in the existing methods. To this end, we combine multi-task learning and distribution learning to address both subjective annotation and intrinsic confusion problems in this paper. The proposed MTAC can adaptively suppress or correct uncertain samples during the modeling training.

\section{Proposed Method}\label{sec:pm}
\begin{figure*}[ht]
   \centering
   \includegraphics[width=1.8\columnwidth]{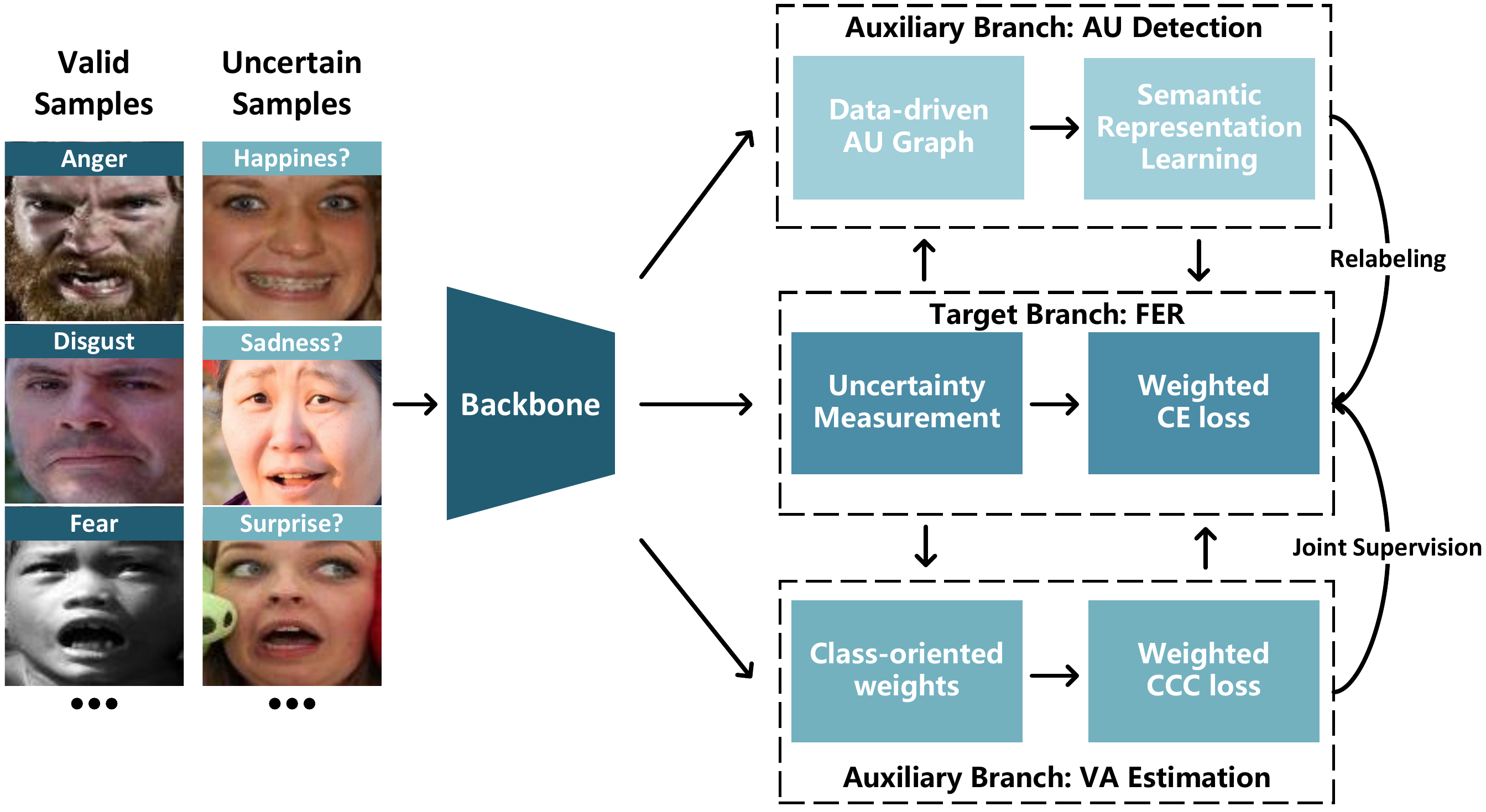}
   \caption{The framework of MTAC. It consists of a target branch for FER, one auxiliary branch for VA estimation, and one auxiliary branch for AU detection. The VA branch supports the feature learning of the target branch by using continuous emotion labels and considering category imbalance, while the AU branch relabels extremely uncertain samples based on the semantic similarity constraint. Samples are re-annotated if they appear semantically closer to a another category center than the original one in feature space. Both auxiliary branches are free to work or disabled and will not participate in the testing process.}\label{fig:framework}
\end{figure*}

As mentioned above, the uncertainty in large-scale FER datasets comes from two aspects, i.e., subjective annotation and intrinsic confusion. To this end, we need to know which samples are uncertain to reduce their impact on model training and correct them to use existing data fully. Inspired by the previous work \cite{chen2020label,zhang2021facial}, we introduce the idea of multi-task learning and distribution learning to achieve the uncertain FER. This work has two bases: 1) features of one sample extracted on similar tasks are correlated, and 2) similar samples should have an underlying dependency in both label space and feature representation. This section presents an overview of the MTAC and then elaborates on its crucial modules.

\subsection{Overview of MTAC}
An overview of the MTAC is illustrated in Fig. \ref{fig:framework}. The MTAC contains: 1) a target branch that takes facial features extracted by a pre-trained backbone network and computes the annotation confidence using a self-attention layer. These confidence weights will affect the importance of the sample when calculating the classification loss. 2) one auxiliary branch of the VA estimation task jointly supervises the feature learning accompanying the class-oriented loss to simultaneously deal with the uncertainty of intrinsic confusion and the category imbalance in the current batch. 3) the other auxiliary branch of the AU detection task constructs data-driven AU graphs, generates a memory template of semantic centers for every emotion category, and then relabels suspicious samples based on the rank regularization and the similarity preserving constraint. The whole MTAC is an end-to-end framework, and the two auxiliary branches can work individually or collaboratively and will not participate in the testing process.

\subsection{Target Branch with Uncertainty Measurement}
\begin{figure*}[ht]
  \centering
  \includegraphics[width=1.8\columnwidth]{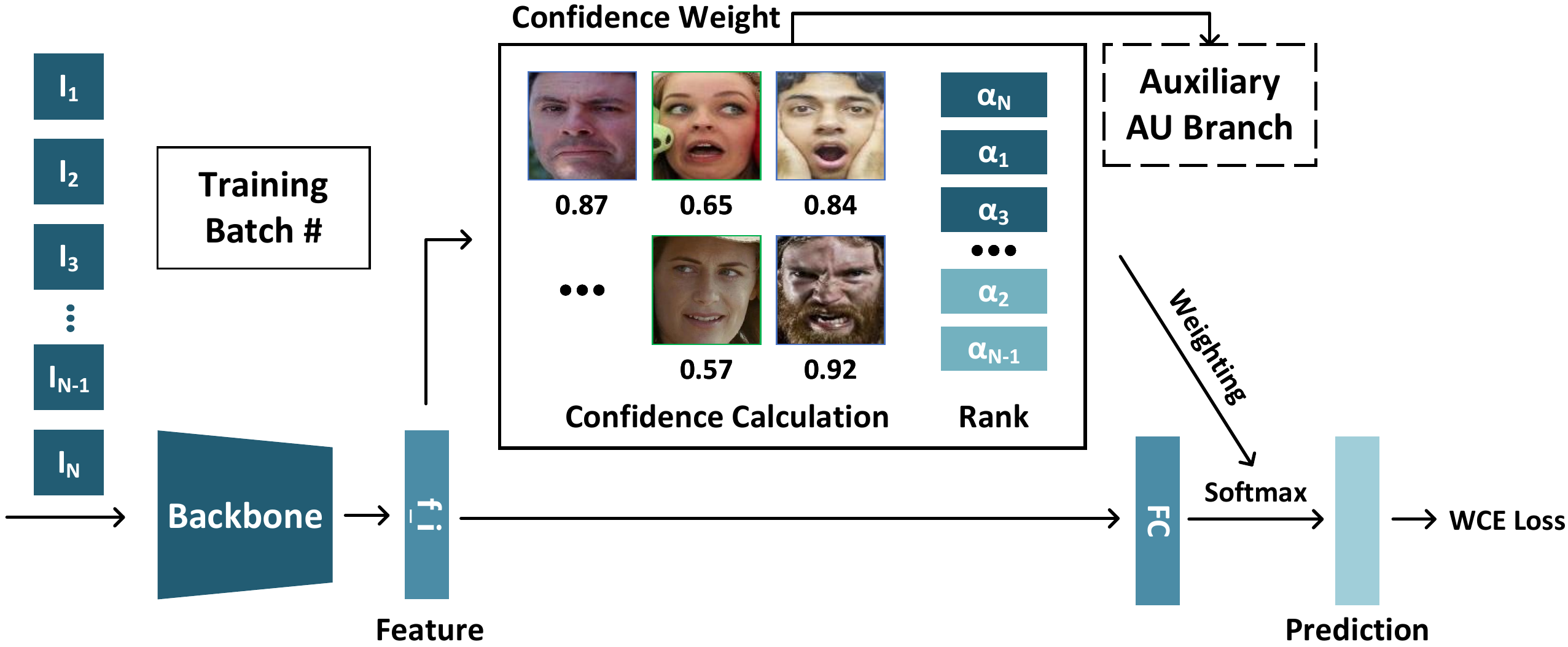}
  \caption{The pipeline of the target branch. Given a training batch, the confidence score of every sample is calculated by applying a self-attention block and is then used to suppress the uncertainty in the loss function. These confidence scores are further passed to the auxiliary AU branch for semantic representation learning and memory template establishment.}\label{fig:target}
\end{figure*}

Before handling the uncertainty, we want the model to provide confidence for each input while making the prediction. As illustrated in Fig. \ref{fig:target}, our target branch follows a general pipeline with a feature extractor and a classifier for the FER task. For a batch of $N$ images, $\bm{F}=[\bm{f}_1,\bm{f}_2,...,\bm{f}_N]\in \mathbb{R}^{D\times N}$ denotes the facial features extracted by the pre-trained backbone network, $D$ indicates the dimension for each facial feature. To identify the ambiguous samples and measure their uncertainties, inspired by \cite{wang2020suppressing, hu2019noise}, a self-attention block is employed that consists of a fully connected (FC) layer and the sigmoid function. Formally, the confidence score of the $i$-th sample can be calculated as: 
\begin{equation}
   \alpha_i = Sigmoid(\bm{W}_a^\top\bm{f}_i), 
\end{equation}
where $\bm{W}_a^\top$ denotes the parameters of the self-attention layer.

During the model training, it is expected that samples with lower confidence should impose less impact, and samples with higher confidence should receive more attention in the current batch. Therefore, we applied the weighted Cross-Entropy (CE) loss similar to \cite{wang2020suppressing, she2021dive}. Specifically, the loss function for the facial expression classifier is formulated as:
\begin{equation}\label{eq:wce}
   L_{wce} = -\frac{1}{N}\sum_{i = 1}^{N}\log\frac{e^{\alpha_i\bm{W}_{y_i}^\top\bm{f}_i}}{\sum_{j = 1}^{C}e^{\alpha_i\bm{W}_j^\top\bm{f}_i}},
\end{equation}
where $\bm{W}_j^\top$ denotes the parameters of the $j$-th classifier, $f_i$ indicates the facial feature, $C$ and $y_i$ are the number of classes and the original discrete label, respectively. According to \cite{liu2017sphereface}, $L_{wce}$ and $\alpha$ are positively correlated.

\subsection{Auxiliary VA Estimation Branch with Category Balancing}\label{sec:va}
\begin{figure}[ht]
  \centering
  \includegraphics[width=0.95\columnwidth]{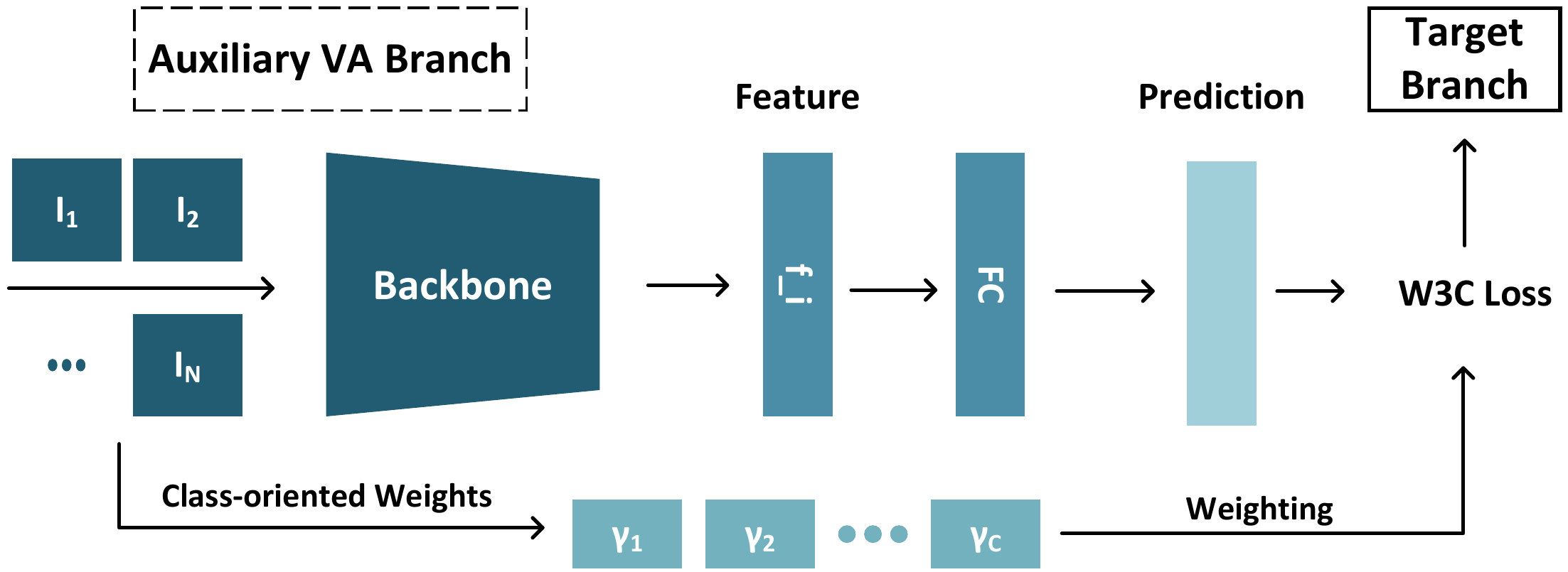}
  \caption{The pipeline of the auxiliary VA branch. Continuous emotion labels are utilized to jointly train the parameter-sharing backbone network for better facial feature learning. In addition, the class-oriented weight is computed to mitigate the category imbalance problem of discrete labels.}\label{fig:va}
\end{figure}

We exploit the VA estimation task as an auxiliary branch to mitigate the uncertainty of intrinsic confusion and complement the biases of discrete emotion labels. As shown in Fig. \ref{fig:va}, the VA estimation branch shares the same backbone network as the target branch. However, it removes the final classifier for continuous predictions of valence and arousal. Specifically, we choose the Concordance Correlation Coefficient (CCC) \cite{tzirakis2017end} as our metric here because it reflects both the trend and the error between the dimensional label and the regressed value, which can be computed as:
\begin{equation}
    \rho = \frac{2\sigma_{y{\hat{y}}}}{\sigma^2_y+\sigma^2_{\hat{y}}+(\mu_y-\mu_{\hat{y}})^2}
\end{equation}
where $y$ and $\hat{y}$ denote the continuous label and the prediction, separately, $\mu$ and $\sigma$ indicate the corresponding mean and variance, respectively, and $\sigma_{y{\hat{y}}}$ is the covariance of $y$ and $\hat{y}$. 

In addition, considering the heavy category imbalance in existing FER datasets, we introduce the class-oriented weight that is designed as:
\begin{equation}
  \gamma_j = 1 - \frac{N_j}{N}, j\in \{1,2,...,C\}, 
\end{equation}
where $N_j$ is the number of images belonging to class $j$, and $C$ is the number of classes. The model should focus more on categories with fewer samples. Class-oriented weight helps to avoid the situation where the training model converges to major classes faster than minor classes. Therefore, we propose a weighted CCC loss function for the VA estimation task as:
\begin{equation}
    L_{w3c} = \sum_{j=1}^{C}\gamma_j(1-\frac{\rho_j^v+\rho_j^a}{2}) 
\end{equation}
where $\rho_j^v$ and $\rho_j^a$ denote the valence CCC and the arousal CCC of the $j$-th category, respectively. We put this category balancing on the VA branch rather than the target branch for two reasons: 1) feature learning in dimensional emotion estimation is not influenced by imbalanced discrete labels; 2) small categories have higher uncertainties of the subjective annotation and larger intra-class distances as shown in Fig. \ref{fig:scatter}. Alternatively, $\gamma$ can also be added to Eq. \ref{eq:wce} similarly to prevent uncertainty from category imbalance when the VA branch is disabled.

\subsection{Auxiliary AU Detection Branch with Graph Reasoning}
\begin{figure}[t]
  \centering
  \includegraphics[width=0.95\columnwidth]{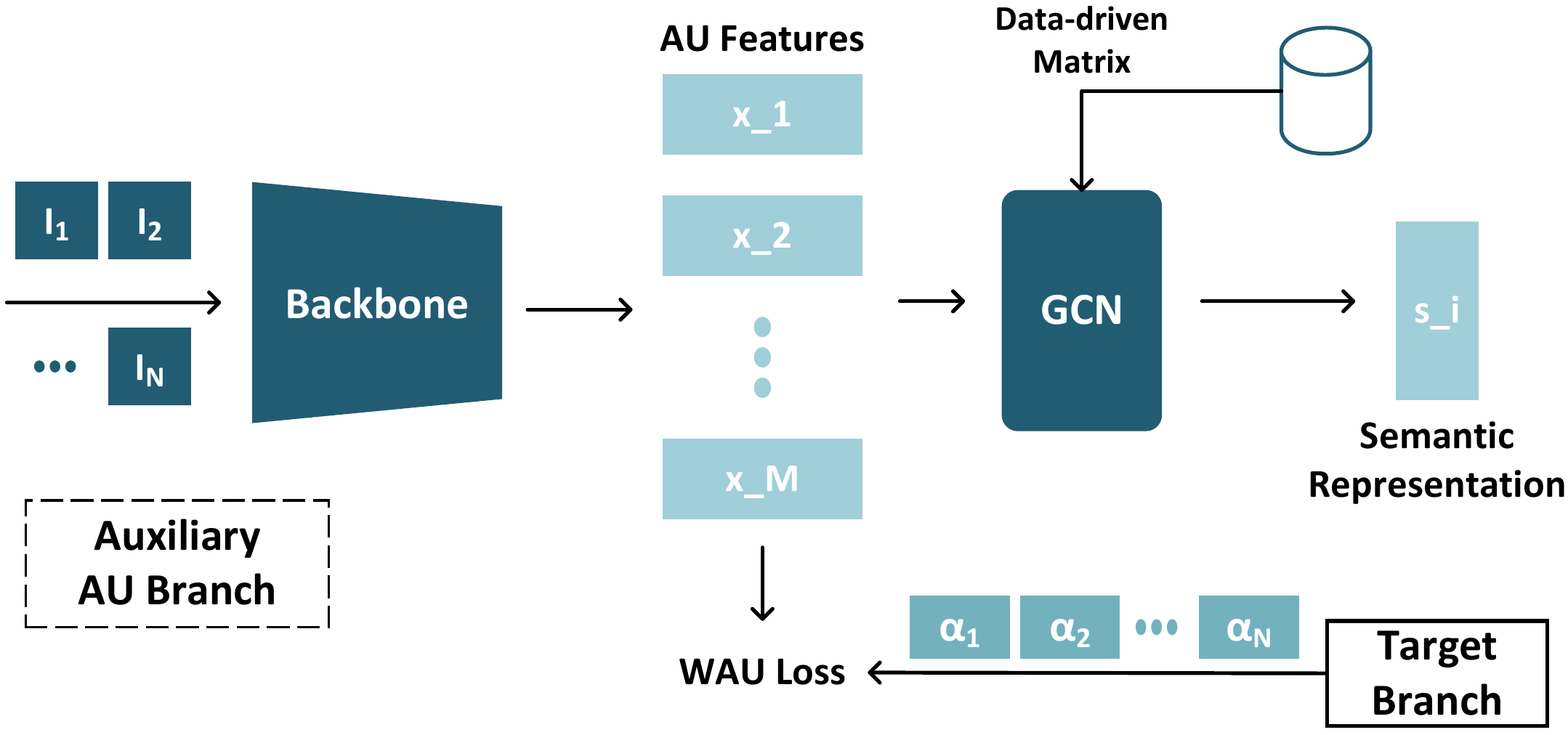}
  \caption{The pipeline of the auxiliary AU branch. The underlying relationship among AUs is encoded by a data-driven graph from datasets and exploited to generate the semantic representation of each sample.}\label{fig:au}
\end{figure}

Although the uncertainty is significantly alleviated with the help of the above two branches, low-confidence samples such as those incorrectly labeled will still degrade the model performance. To this end, the AU detection task is further employed as the other auxiliary branch because the Facial Action Coding System is proven to have latent mappings with emotion categories \cite{friesen1978facial,liu2019facial,liu2022graph}. As illustrated in Fig. \ref{fig:au}, we can obtain a set of AU features of each image through the backbone network, $\bm{X}^i=[\bm{x}_1^i,\bm{x}_2^i,...,\bm{x}_M^i]\in\mathbb{R}^{B\times M}$, $B$ and $M$ denote the feature dimension and the AU amount, respectively. 

Considering the consistency of predefined mappings between emotion categories and AUs in large-scale datasets is difficult to guarantee \cite{liu2021sg,lei2021micro}, we construct a data-driven AU graph that takes individual AU features as graph nodes and the co-occurring AU dependency as graph edges. Specifically, our AU graph is based on the conditional probability of obtaining co-occurring AU dependencies from the training set, which can be calculated as:
\begin{equation}
	\bm{A}_{p,q} = P(AU_p|AU_q) = \frac{OCC_{p\cap q}}{OCC_q},
\end{equation}
where $OCC_{p\cap q}$ denotes the number of co-occurrences of $AU_p$ and $AU_q$, and $OCC_q$ is the total number of occurrences of $AU_q$. Since the AU co-occurring relationship is practically asymmetric, so $P(AU_p|AU_q) \neq P(AU_q|AU_p)$.

Then, the AU graph is input in a two-layer GCN to extract the semantic representation. Formally, each graph convolution layer is represented as: 
\begin{equation}
   \bm{X}'=g(\bm{X},\bm{A})=LeakyRELU(\bar{\bm{A}}\bm{X}\bm{W}_g),
\end{equation}
where $\bar{\bm{A}}$ denotes the normalized $\bm{A}$ with all rows sum to one, $\bm{W}_g$ is the weight matrix to be learned in the current layer.

All the node features outputted by the GCN are fed into a FC layer with Sigmoid functions to predict multiple AUs. Similar to $L_{wce}$, we improve the binary CE loss with the confidence score to train every AU classifier, and the total weighted group loss for the AU branch is formulated as:
\begin{equation}
   L_{wau} = -\sum_{m = 1}^{M}{\alpha(z_m\log{p_m} + (1-z_m)\log{(1-p_m)})}, 
\end{equation}
where $\alpha$ is the confidence weight, $z_m$ and $p_m$ are the original/pseudo label and the prediction of $m$-th AU, respectively. The logits $\bm{s}_i\in \mathbb{R}^{1\times M}$ before AU classifiers are treated as the semantic representation of the sample. 

\subsection{Relabeling with Semantic Similarity Constraint}
\begin{figure}[t]
  \centering
  \includegraphics[width=0.95\columnwidth]{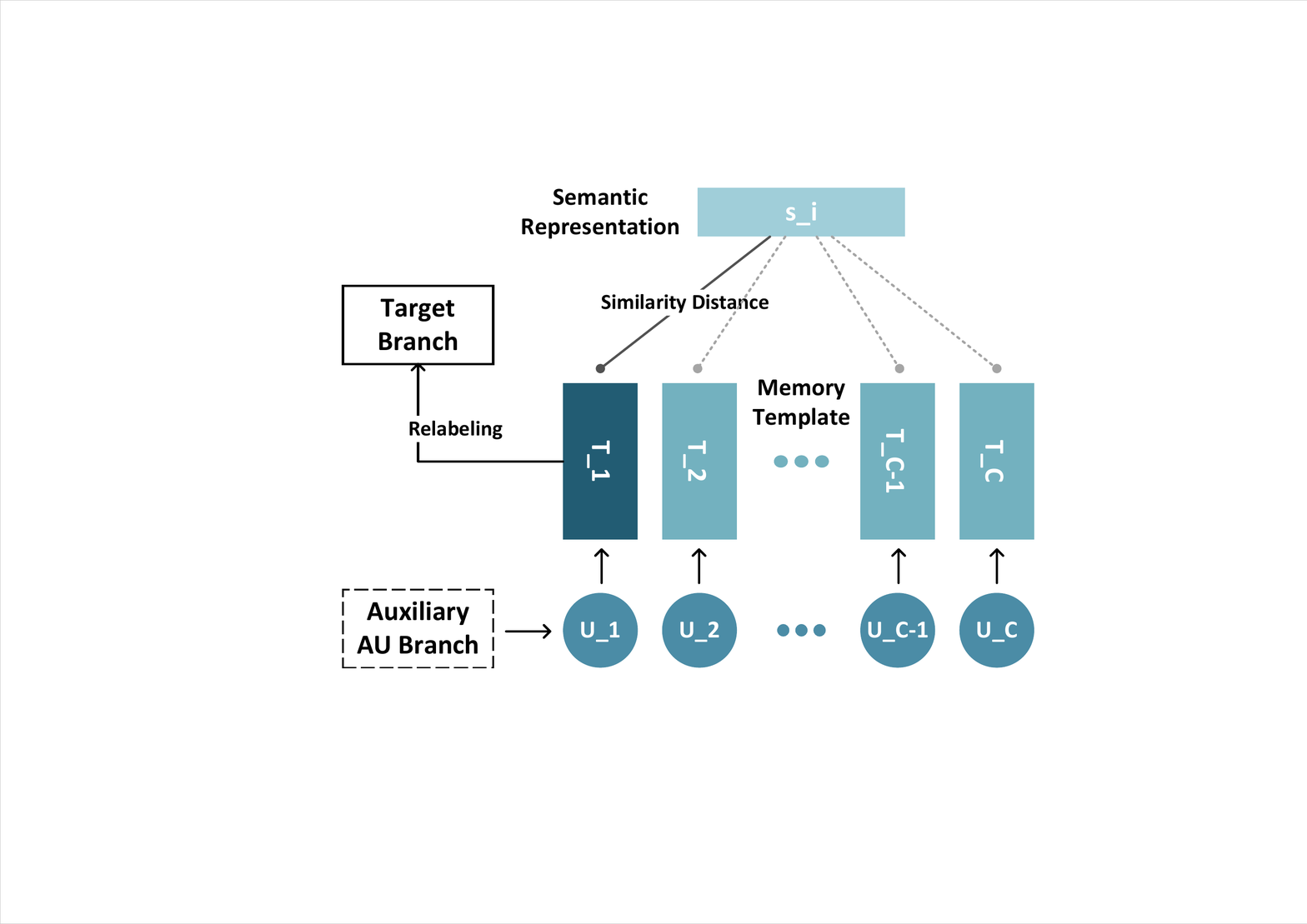}
  \caption{The pipeline of the relabeling strategy. A memory template is built and updated based on the average category centers and then constrained the relabeling strategy with similarity distance. The new label then participates in network optimization in the target branch.}\label{fig:relabeling}
\end{figure}

To determine which labels need to be corrected and which new classes should be assigned, we design a semantic similarity constrained relabeling strategy (see Fig. \ref{fig:relabeling}). For every training batch, a center set for all emotion categories $\bm{U}=[\bm{u}_1,\bm{u}_2,...,\bm{u}_C]\in\mathbb{R}^{M\times C}$ is generated based on the semantic representations and the confidence weights, which can be calculated as:
\begin{equation}
  \bm{U}_j=\frac{1}{N_j}\sum_{n_j=1}^{N_j}\alpha_{n_j}\bm{s}_{n_j},
\end{equation}
where $N_j$ is the number of the samples with the $j$-th label in the current batch. Then, a memory template $\bm{T}\in\mathbb{R}^{M\times C}$ is initialized and dynamically updated throughout the whole training process as follows:
\begin{equation}
    \bm{T}_j = (1-e^{-\tau h})\bm{T}_j + e^{-\tau h}\bm{U}_j, \tau \in (0,1]
\end{equation}
where $h$ denotes the batch index, and $\tau$ is a control factor of updating rate. Eventually, the memory template will gradually stabilize as the model converges \cite{cheng2022entropy}. After that, the cosine distance between each semantic representation $\bm{s}_i$ and each of semantic center $\bm{t}_j$ in the memory template $\bm{T}$ is computed as:
\begin{equation}
  dist(\bm{s}_i,\bm{t}_j)=1-\frac{\bm{t}_j \cdot \bm{s}_i}{\Vert \bm{t}_j\Vert \Vert \bm{s}_i\Vert}.
\end{equation}

Next, for every sample in the current batch, we rank all its semantic distances to the memory template $T$. Benefiting from the other two branches, it is supposed that the uncertain samples should be suppressed and have a large distance from their original category center. Thus, for those samples with extreme uncertainty, we relabel them following the semantic similarity constraint, which can be defined as:
\begin{equation}
  y'_i=\begin{cases}
    j, & \mbox{if } dist(\bm{s}_l,\bm{t}_{org}) > min(dist(\bm{s}_l,\bm{t}_j)) \\
    y_i, & \mbox{otherwise}  
    \end{cases}
\end{equation}
where $y'_i$ denotes the corrected label, $org$ indicates the original discrete category, and $j\neq org$. Note that this relabeling strategy will only take effect if the semantic distance to the original category center is not the shortest in the ranking. In such cases, the template class with the highest semantic similarity will be assigned to this sample as a new label. 

\subsection{Model Training}
Finally, the total loss function of the whole framework can be formulated as:
\begin{equation}
  L_{total}= \lambda_1(L_{wce}+L_{w3c})+\lambda_2 L_{wau},
\end{equation}
where $\lambda_1$ and $\lambda_2$ are the weighted ramp functions that will change with epoch rounds \cite{laine2016temporal}, which can be computed as follows:
\begin{equation}
	\lambda_1 = \begin{cases}
		\exp(-(1-\frac{\beta}{H})^2), & \beta\leq H\\
		1, & \beta > H
	\end{cases},
\end{equation}
\begin{equation}
	\lambda_2 = \begin{cases}
		1, & \beta \leq H \\
		\exp(-(1-\frac{H}{\beta})^2), & \beta > H
	\end{cases},
\end{equation}
where $\beta$ denotes the current epoch index, and $H$ is a constant that controls the participation of different branches. 

The weighted ramp functions allow MTAC to pay more attention to the AU branch in the initial training stage. Since the number of samples accumulated at the beginning is insufficient, it cannot generate effective semantic representations and solid memory templates. After a certain number of training rounds, the model will focus more on the target and VA branches to extract distinguishing features for final predictions. Moreover, our MTAC can work independently with the target branch, while the two auxiliary branches can be flexibly combined into the framework without additional inference burden. 

\section{Experiments}\label{sec:exp}
In this section, we conduct extensive experiments to demonstrate the performance of MTAC in terms of ablation study, tackling synthetic and real uncertainty, and multi-task comparison with the state-of-the-art. 

\subsection{Datasets}
Three challenging FER benchmarks are used for evaluation, i.e., RAF-DB \cite{li2019reliable}, AffectNet \cite{mollahosseini2017affectnet}, and AffWild2 \cite{kollias2022abaw}. All three datasets have unconstrained conditions and large-scale samples, containing subjective or intrinsic uncertainty. 

\textbf{RAF-DB} is a large-scale dataset with $15,339$ in-the-wild images annotating of six basic emotions and neutral. In our experiments, $12,271$ and $3,368$ samples are used for training and testing, respectively. Since no continuous emotion labels are provided, the VA branch will be disabled in the experiment on RAF-DB. 

\textbf{AffectNet} contains close to one million expression images. To ensure a fair comparison, we select samples manually labeled as six basic emotions and neutral for evaluation. The number of images in the training and test sets is $283,901$ and $3,500$, respectively. In addition, automatically labeled samples in AffectNet are used as a set of real noisy data, denoted as \textbf{AffectNet\_Auto}, to verify the ability of MTAC to handle uncertain expressions.

\textbf{AffWild2} is the first audiovisual dataset with annotations for all main behavior tasks, including FER, VA estimation, and AU detection. It contains 558 videos with around 2.8 million images of facial expressions. In this work, we use the subset of 'MTL\_Challenge' with seven discrete labels, VA labels, and AU labels simultaneously, which consists of $39,614$ and $10,839$ training and testing samples, respectively. The CCC score is used as a metric for the evaluation of the regression task.

Since the AU annotation requires specially trained experts and is time-consuming, it is natural that no AU labels are provided in RAF-DB and AffectNet. To account for this issue, we applied Openface 2.0 \cite{baltrusaitis2018openface} to automatically generate pseudo AU labels, similar to \cite{chen2020label, chen2021understanding}. For AffWild2, the original AU labels are used to generate the AU graph. In other words, the AU branch and the relabeling can work either fully or weakly supervised and compatible with various datasets. Moreover, the AU branch does not participate in the parameter update of the backbone network. Our MTAC utilizes a feature-level semantic similarity constraint to correct the extremely uncertain sample adaptively instead of directly replacing it with the prediction, which can reduce the negative impact of incorrect pseudo AU labels. 

\subsection{Implementation Details}
The MTAC is implemented with the Pytorch platform and trained using two Nvidia Volta V100 GPUs. The cropped facial regions are used and further resized to $224\times 224$ pixels as inputs. For the whole framework, we choose the ResNet-18 \cite{he2016deep} and the DenseNet \cite{huang2017densely} as two backbone networks which are pre-trained on the MS-Celeb-1M \cite{guo2016ms} dataset as previous methods \cite{she2021dive, wang2020suppressing,wang2020region}. For the target branch and the auxiliary VA branch, the initial learning rate of the Adam optimizer is $0.01$, which is then updated to $10^{-3}$ and $10^{-4}$ at the $10$-th and $20$-th epoch, respectively. For the auxiliary AU branch, each GCN layer has 64 channels, where the control factor $\tau$ and the decayed learning rate are set as $0.9$ and $0.005$, respectively. We choose a batch size $512$ to ensure that every template can be effectively updated during the whole training process, while the $H$ defaults to $5$. The relabeling starts after $10$ epochs to ensure a stable memory template of the semantic representation. 

\subsection{Ablation Study}
A few ablation studies are performed to verify the contribution of every branch in MTAC and the key hyper-parameter proposed in this paper.

\subsubsection{Components evaluation}
MTAC deals with the effects of uncertain samples based on three branches, i.e., target branch, auxiliary VA branch, and auxiliary AU branch. The target branch suppresses suspicious samples and highlights valid inputs through confidence measurement and weighted loss function. The VA branch optimizes the parameter-sharing network with continuous annotations and considers the category imbalance. The AU branch corrects extremely uncertain labels with the data-driven AU graph and semantic memory templates. All three branches can be flexibly combined with various network architectures. In this experiment, we design five different settings for effectiveness verification. Note that the class-oriented weights are assigned in the target branch on RAF-DB due to the lack of continuous labels. When no branch is active, it is equivalent to a standard ResNet-18. 

As shown in Tab. \ref{tab:comp}, the independent target branch significantly enhances the FER performance on three datasets, and a greater improvement can be achieved by further using two auxiliary branches, respectively. In particular, the VA branch performs slightly better than the AU branch on AffectNet and AffWild2 because of the additional knowledge from the continuous label space and the manipulation for the huge category imbalance. The best performance comes from the complete MTAC framework with all three branches that consider the uncertainty from both subjective annotation and intrinsic confusion.

\begin{table}[h]
   \centering
   \caption{Evaluation of different branches. '\emph{Target B.(ranch)}' applies the uncertainty measurement, auxiliary \emph{VA B.(ranch)} executes the joint feature learning and the category balancing, and auxiliary \emph{AU B.(ranch)} exploits the data-driven AU graph and the semantic similarity constrained relabeling. \textbf{Bold} denotes the best result, and \textit{ITALICS} indicates the second best result.}
   \label{tab:comp}
   \setlength{\tabcolsep}{2.5mm}{
   \begin{tabular}{c|c|c|c|c|c}
     \hline
     \textit{Target B.} & \textit{VA B.} & \textit{AU B.} & RAF-DB & AffectNet & AffWild2\\
     \hline
     $\times$ & $\times$ & $\times$ & 85.81 & 57.94 & 56.41 \\
     \checkmark & $\times$ & $\times$ & \textit{86.55} & 61.97 & 59.09\\
     \checkmark & \checkmark & $\times$ & - & \textit{63.71} & \textit{61.47} \\
     \checkmark & $\times$ & \checkmark & \textbf{89.31} & 62.51 & 61.16 \\
     \checkmark & \checkmark & \checkmark & - & \textbf{65.09} & \textbf{62.78} \\
     \hline
   \end{tabular}}
\end{table}

\subsubsection{Evaluation of the class-oriented weight}
Most large-scale facial expression datasets have severe category imbalances. In MTAC, the proposed class-oriented weight $\gamma$ is compatible with various loss functions. In this experiment, we design three different settings, i.e., MTAC without $\gamma$, $\gamma$ in the target branch (as our preliminary work in ICPR 2022 \cite{liu2022uncertain}), and $\gamma$ in the auxiliary VA branch. As presented in Tab. \ref{tab:balance}, the category balancing significantly contributes to the model training. It shows a better performance in the VA branch, which demonstrates our statement in Sec. \ref{sec:va}.

\begin{table}[h]
	\centering
	\caption{Evaluation of the class-oriented weight. \textbf{Bold} denotes the best result, and \textit{ITALICS} indicates the second best result.}\label{tab:balance}
	\label{tab:class}
	\setlength{\tabcolsep}{4mm}{
		\begin{tabular}{c|c|c|c}
			\hline
			Method & RAF-DB & AffectNet & AffWild2\\
			\hline
			w/o $\gamma$ & 88.45 & 63.46 & 61.83 \\
            \textit{Target B.} w/ $\gamma$ & \textbf{89.31} & \textit{64.83} & \textit{62.07} \\
            \textit{VA B.} w/ $\gamma$ & - & \textbf{65.09} & \textbf{62.78} \\
			\hline
	\end{tabular}}
\end{table}

\subsubsection{Evaluation of the data-driven graph}
Relabeling under semantic similarity constraints is another essential module of MTAC for uncertainty mitigation. Its semantic information of AU co-occurring dependencies is encoded with a data-driven AU graph. To study the established edges, we randomly initialize $A$ with element values from $0$ to $1$ to shield edge attributes in this experiment. We also design a fully-connected $A$ that every element is fixed as $1$. As shown in Tab. \ref{tab:edge}, the random edges introduce additional uncertainty and lead to performance decreases, while the fixed edges can not reflect the AU co-occurrence and approximate the actual distribution. On the contrary, our data-driven AU graph helps the GCN to generate better semantic representations and further boosts the memory templates and the relabeling.

\begin{table}[h]
	\centering
	\caption{Evaluation of the data-driven graph. \textbf{Bold} denotes the best result, and \textit{ITALICS} indicates the second best result.}
	\label{tab:edge}
	\setlength{\tabcolsep}{3mm}{
		\begin{tabular}{c|c|c|c}
			\hline
			Method & RAF-DB & AffectNet & AffWild2 \\
			\hline
			w/ \emph{random edges} & 86.10 & 62.06 & 60.88 \\
            w/ \emph{fixed edges} & \textit{87.14} & \textit{63.94} & \textit{61.63} \\
			w/ \emph{data-driven edges} & \textbf{89.31} & \textbf{65.09} & \textbf{62.78} \\
			\hline
	\end{tabular}}
\end{table}

\subsection{Evaluation of Handling Uncertainty}
To test our MTAC in handling uncertain samples, we set up comparative experiments under synthetic uncertainty and real uncertainty, respectively. Two baseline methods, i.e., ResNet-18 and DenseNet, and two state-of-the-art methods with uncertainty consideration, i.e., SCN \cite{wang2020suppressing}, and DMUE \cite{she2021dive}, are selected for comparison.

\subsubsection{Synthetic uncertain samples}
We randomized 10\%, 20\%, and 30\% of the original labels of the training set for RAF-DB and AffectNet, respectively. From Tab. \ref{tab:syn}, the proposed MTAC outperforms baselines on two datasets, illustrating that uncertain samples hamper network training. Moreover, as the proportion of uncertainty increases, the performance degradation of MTAC compared to the corresponding baselines is also smaller, which further proves the effectiveness of our feature-level semantic similarity constraint. The DMUE achieves the best results on AffectNet by multi-branch distribution learning when facing 10\% and 20\% uncertainty. Benefiting from multi-task correction, our MTAC obtains competitive performance in the above two settings and performs the best in the experiment with 30\% uncertainty.

\begin{table}[h]
   \centering
   \caption{Evaluation of encountering synthetic uncertain samples. \textbf{Bold} denotes the best result, and \textit{ITALICS} indicates the second best result. $*$ means performing 8-category classification.}\label{tab:syn}
   \setlength{\tabcolsep}{3mm}{
   \begin{tabular}{c|c|c|c}
     \hline
     Method & Uncertainty & RAF-DB & AffectNet$^*$ \\
     \hline
     ResNet-18 & \multirow{6}*{10\%} & 80.64 & 57.25 \\
     DenseNet & & 81.03 & 57.50 \\
     SCN \cite{wang2020suppressing} &  & 82.18 & 58.58 \\
     DMUE \cite{she2021dive} &  & 83.19 & \textbf{61.21} \\
     MTAC (ResNet-18) &  & \textit{83.22} & 59.45 \\
     MTAC (DenseNet) &  & \textbf{83.64} & \textit{60.20} \\
     \hline
     ResNet-18 & \multirow{6}*{20\%} & 78.06 & 56.23 \\
     DenseNet &  & 79.48 & 56.98 \\
     SCN \cite{wang2020suppressing} &  & 80.10 & 57.25 \\
     DMUE \cite{she2021dive} &  & 81.02 & \textbf{59.06} \\
     MTAC (ResNet-18) &  & \textit{81.15} & 58.50 \\
     MTAC (DenseNet) &  & \textbf{81.92} & \textit{59.05} \\
     \hline
     ResNet-18 & \multirow{6}*{30\%} & 75.12 & 52.60 \\
     DenseNet &  & 76.51 & 52.80 \\
     SCN \cite{wang2020suppressing} &  & 77.46 & 55.05 \\
     DMUE \cite{she2021dive} &  & \textit{79.41} & \textit{56.88} \\
     MTAC (ResNet-18) &  & 79.01 & 56.45 \\
     MTAC (DenseNet) &  & \textbf{80.86} & \textbf{57.33} \\
     \hline
   \end{tabular}}
\end{table}

\subsubsection{Real uncertain samples}
Apart from synthetic uncertainty, we also select AffectNet\_Auto as a training set with naturally uncertain samples of wrong annotations and confusing emotions for cross-dataset validation, which is rarely considered by previous studies. The automatic labeling algorithm published in the official document has an accuracy of $65\%$ \cite{mollahosseini2017affectnet}. As shown in Tab. \ref{tab:real}, MTAC achieves the best when encountering real uncertain samples, and the performance growth exceeds that in the synthetic uncertainty experiment. One possible explanation is that we additionally accounted for the intrinsic confusions, which are more general uncertainty in real data. In the proposed MTAC, the class-oriented weight in the auxiliary VA branch mitigates ambiguities from imbalanced categories, and the semantic memory template with updated category centers in the auxiliary AU branch conducts effective label correction.

\begin{table}[h]
	\centering
	\caption{Performance of MTAC on datasets with real uncertain samples. \textbf{Bold} denotes the best result, and \textit{ITALICS} indicates the second best result. $^\dagger$ means re-implementing results.}
	\label{tab:real}
	\setlength{\tabcolsep}{10mm}{
	\begin{tabular}{c|c}
    \hline
    Method & AffectNet\_Auto \\
    \hline
    ResNet-18 & 53.23 \\
    DenseNet & 53.83 \\
    SCN$^\dagger$ \cite{wang2020suppressing} & 55.43 \\
    DMUE$^\dagger$ \cite{she2021dive} & 56.98 \\
    MTAC (ResNet-18) & \textit{57.38} \\
    MTAC (DenseNet) & \textbf{57.80} \\
    \hline
	\end{tabular}}
\end{table}

\subsection{Visualization}
To present the specific manipulation effect of MTAC on uncertain samples, we visualize the intermediate results in terms of passive uncertainty suppression with uncertainty measurement and active uncertainty correction with relabeling. 

\subsubsection{Uncertainty measurement}
\begin{figure*}[h]
  \centering
  \includegraphics[width=1.8\columnwidth]{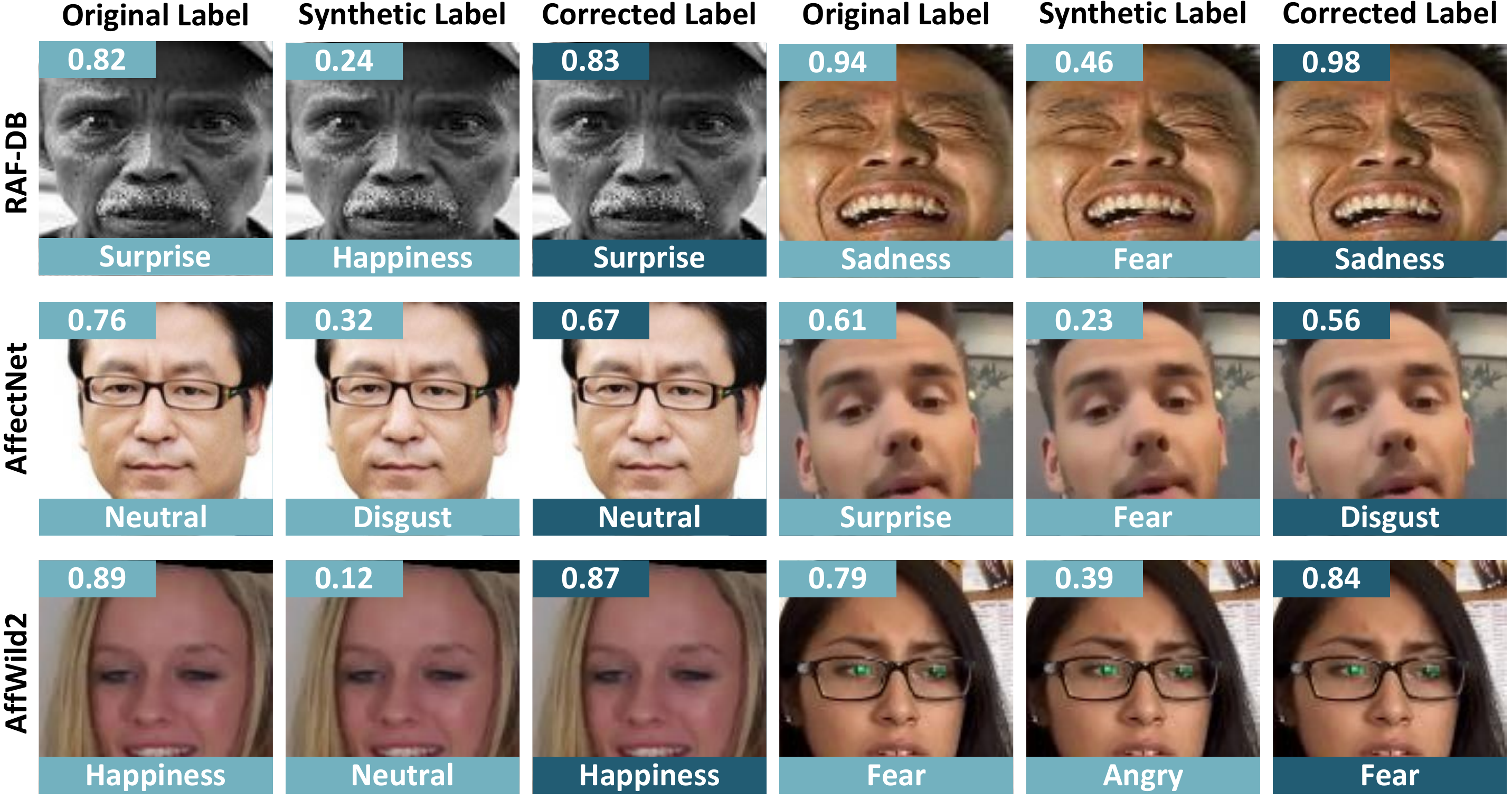}
  \caption{Visualization of joint feature learning. Two examples of each dataset in RAF-DB, AffectNet, and AffWild2 are shown. The top left block denotes the confidence score $\alpha$, and the bottom block presents the label of the current sample. From left to right of every three columns are the original sample, synthetic sample, and corrected sample, respectively. }\label{fig:vis-tar}
\end{figure*}

Fig. \ref{fig:vis-tar} depicts the visualization of the uncertainty measurement in the target branch on examples in RAF-DB, AffectNet, and AffWild2. Generally, the proposed MTAC successfully figures out the uncertain samples based on the confidence score and adaptively updates the value after the relabeling execution. In particular, in the second case of AffectNet, our MTAC accurately identifies the synthetic uncertainty and performs a correction for the original annotation. 

\subsubsection{Relabeling}
\begin{figure*}[h]
  \centering
  \includegraphics[width=1.95\columnwidth]{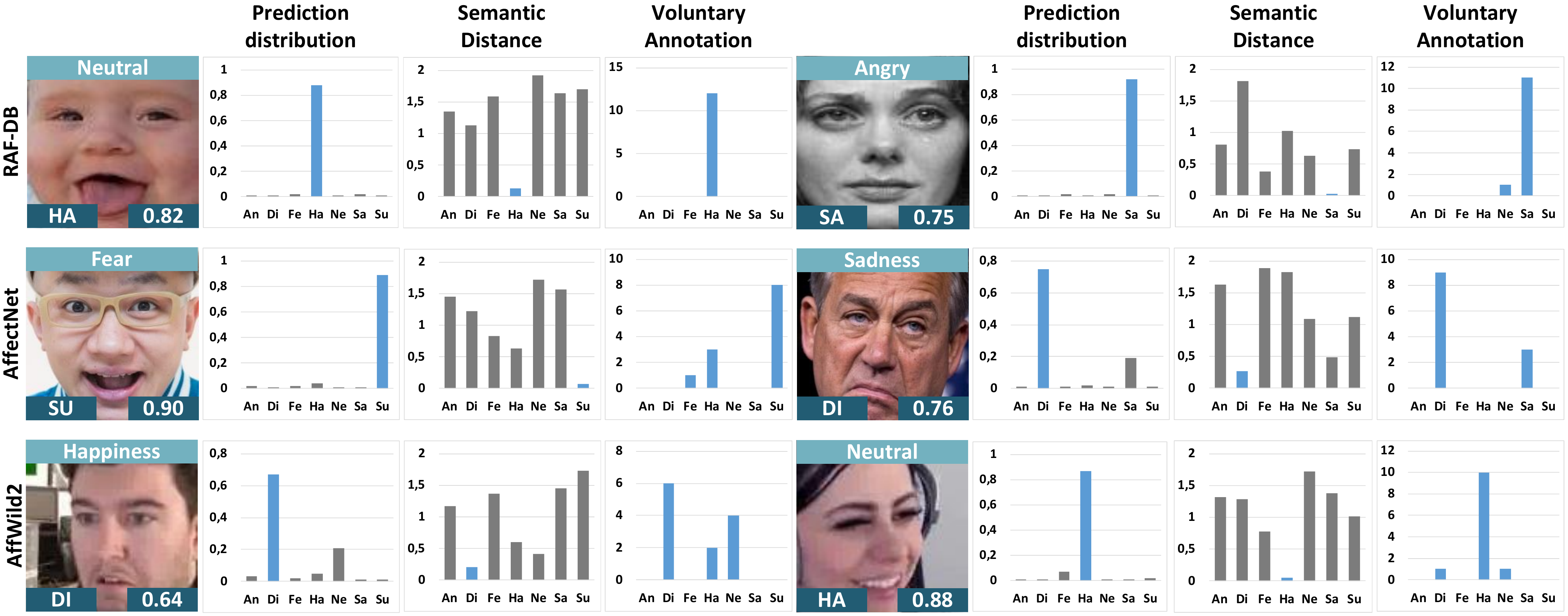}
  \caption{Visualization of relabeling. Two examples of each dataset in RAF-DB, AffectNet, and AffWild2 are shown. The light color block at the top denotes the synthetic uncertain label, the dark color block at the bottom left indicates the new label after relabeling, and the dark block at the bottom right presents the confidence score $\alpha$ after correction. From left to right of every four columns are the original sample, prediction distribution, semantic similarity distance, and voluntary annotation statistic, respectively. DI, HA, SA, and SU are \emph{disgust}, \emph{happiness}, \emph{sadness}, and \emph{surprise}, respectively.}\label{fig:vis-aux}
\end{figure*}

To exhibit the semantic similarity constrained relabeling workflow, we illustrate the prediction distribution in the target branch and the semantic distance in the auxiliary AU branch on examples in RAF-DB, AffectNet, and AffWild2. In addition, subjective annotations from twelve volunteers are counted to make a comparison with our relabeling strategy. As shown in Fig. \ref{fig:vis-aux}, the generated memory template of semantic representation centers can increase the inter-class distance. The predicted emotion categories are similar in distribution to manual annotations. It reveals that our MTAC can effectively handle uncertain samples to facilitate the model training and improve the final FER performance.

\subsection{Comparison with the State-of-the-art}
Since the proposed MTAC is designed for FER on large-scale datasets and utilizes multiple labels, we compare it with the state-of-the-art methods for single-task and multi-task performance evaluation.

\subsubsection{Evaluation of single FER task}
\begin{figure*}[h]
	\centering
    \subfloat[RAF-DB]{\includegraphics[width=0.66\columnwidth]{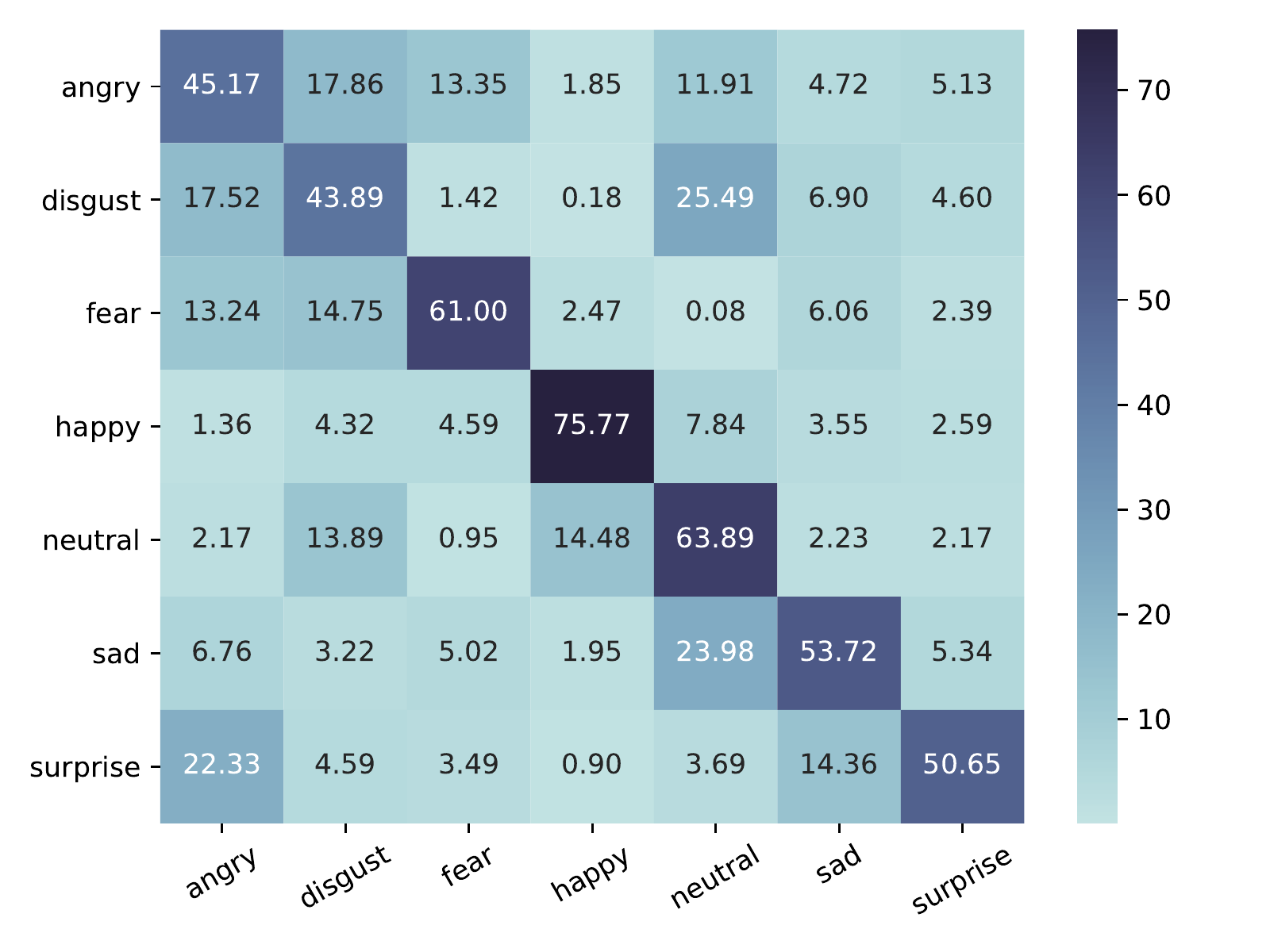}}
    \subfloat[AffectNet]{\includegraphics[width=0.66\columnwidth]{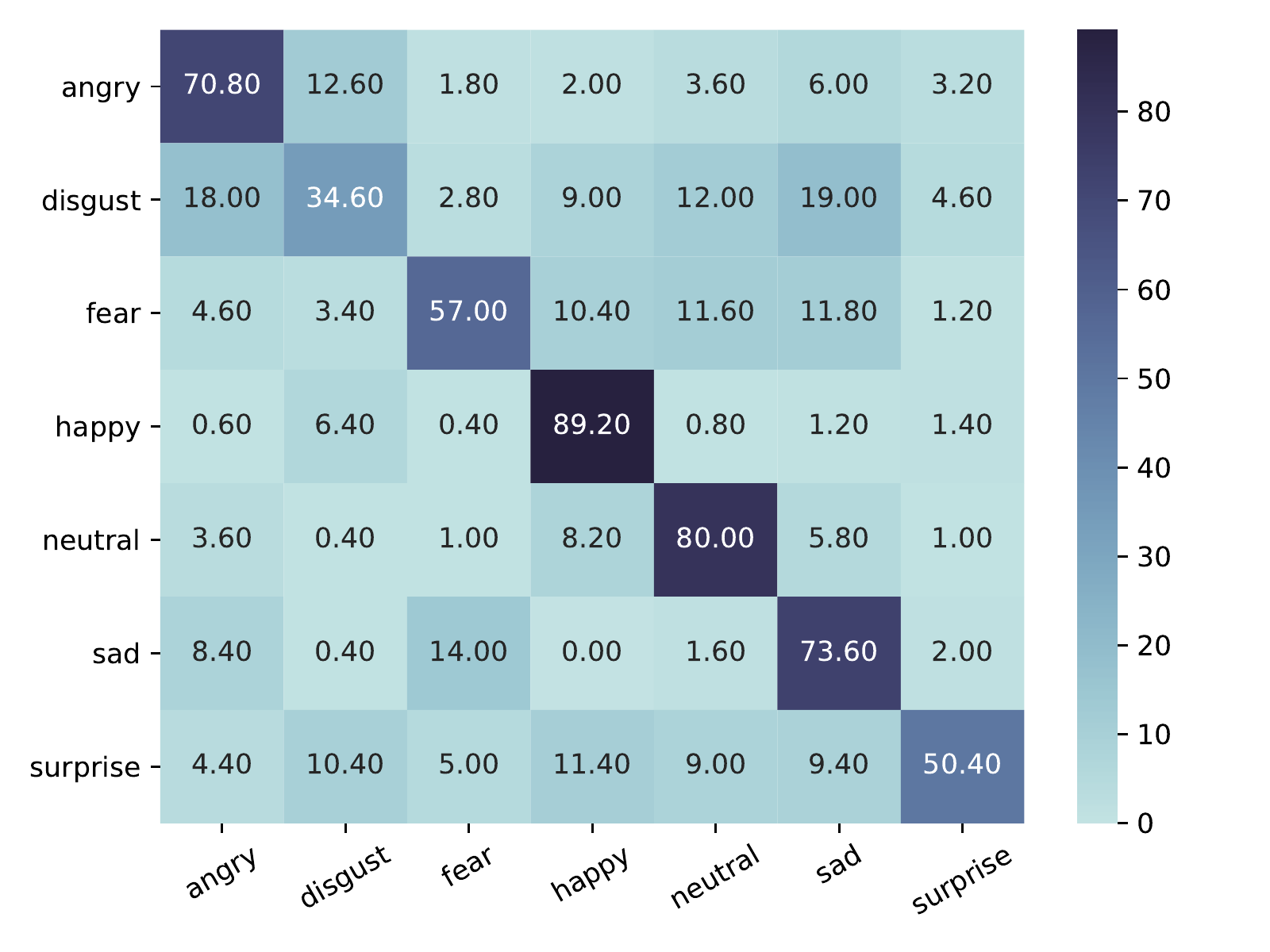}}
    \subfloat[AffWild2]{\includegraphics[width=0.66\columnwidth]{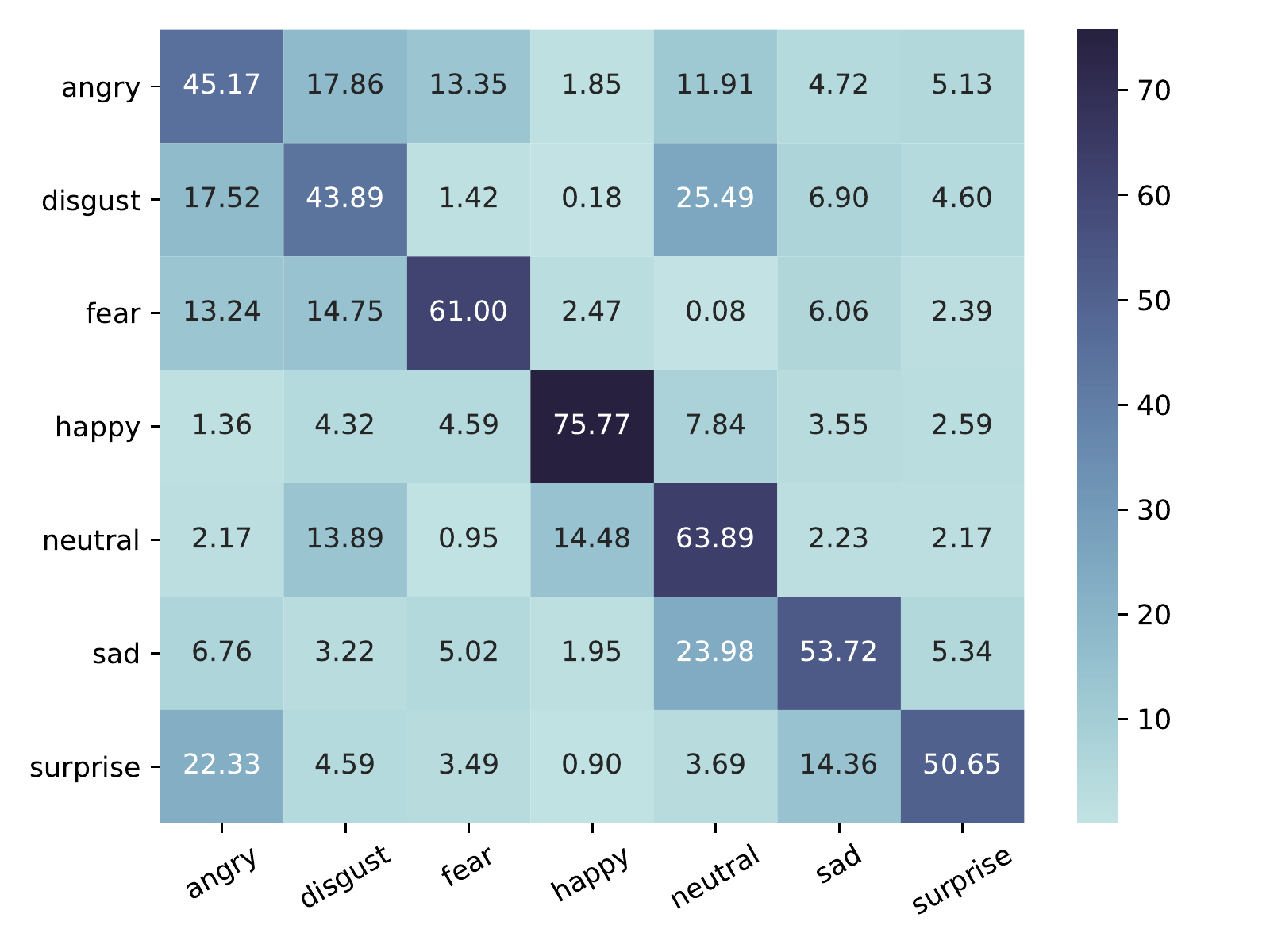}}
	\caption{Confusion matrices of MTAC. }\label{fig:cm}
\end{figure*}

Tab. \ref{tab:soat} shows the performance comparison and Fig. \ref{fig:cm} presents the confusion matrices of MTAC in the single FER task. To summarize, our method performs the best and the top-2 results on RAF-DB and AffectNet, respectively. Although LDL-ALSG \cite{zeng2018facial}, SEIIL \cite{li2021self}, and Face2Exp \cite{zeng2022face2exp} introduce extra knowledge to support the network training, they only consider the label-level distribution and cannot repair the uncertain samples, leading to performance degrade. In addition, IPA2LT \cite{zeng2018facial}, SCN \cite{wang2020suppressing}, WSND \cite{zhang2021weakly}, and FENN explicitly deal with uncertain labels and thus achieve good results. However, intrinsic uncertainty can still limit their feature learning without information in the side space. Benefiting from the uncertainty measurement, the data-driven AU graph, and the feature-level constrained relabeling, our MTAC outperforms NMA \cite{zhang2021facial} and achieves competitive results against DMUE \cite{she2021dive} that apply uncertainty mitigation and auxiliary task simultaneously, which reveal the effectiveness of the proposed modules in this work. Note that 

\begin{table}[h]
   \centering
   \caption{Comparisons with the state-of-the-art methods. $*$ means performing 8-category classification. $\dagger$ indicates the uncertainty handling is introduced. $\ddagger$ denotes extra knowledge of auxiliary tasks is considered. \textbf{Bold} denotes the best result, and \textit{ITALICS} indicates the second best result. }\label{tab:soat}
   \setlength{\tabcolsep}{4mm}{
   \begin{tabular}{c|c|c|c}
     \hline
     Method & Year & RAF-DB & AffectNet$^*$ \\
     \hline
     IPA2LT$^\dagger $\cite{zeng2018facial} & 2018 & 86.77 & 55.11 \\
     SCN$^\dagger $ \cite{wang2020suppressing} & 2020 & 88.14 & 60.23\\
     RAN \cite{wang2020region} & 2020 & 86.90 & 59.50 \\
     LDL-ALSG$^\ddagger$ \cite{chen2020label} & 2020 & 85.53 & 59.35 \\
     SPWFA-SE \cite{li2020facial} & 2020 & 86.31 & 59.23 \\
     SEIIL$^\ddagger$ \cite{li2021self} & 2021 & 88.23 & - \\
     WSND$^\dagger$ \cite{zhang2021weakly} & 2021 & 88.89 & 60.04 \\
     NMA$^{\dagger\ddagger}$ \cite{zhang2021facial} & 2021 & 76.10 & 46.08 \\
     DMUE$^{\dagger\ddagger}$ \cite{she2021dive} & 2021 & 88.76 & \textbf{63.11} \\
     IDFL \cite{li2022learning} & 2022 & 86.96 & 59.20 \\
     Face2Exp$^{\ddagger}$ \cite{zeng2022face2exp} & 2022 & 88.54 & - \\
     FENN$^{\dagger}$ \cite{gu2022towards} & 2022 & \textit{88.91} & 60.83 \\
     \hline
     MTAC (ResNet-18)$^{\dagger\ddagger} $ & Ours & 88.45 & 61.58 \\
     MTAC (DenseNet)$^{\dagger\ddagger} $ & Ours & \textbf{89.31} & \textit{61.90} \\
     \hline
   \end{tabular}}
\end{table}

\subsubsection{Multi-task evaluation}
To demonstrate the capability of MTAC in executing multi-task prediction, we introduce two advanced approaches re-implemented by ourselves for further evaluation, i.e., Emotion-GCN \cite{antoniadis2021exploiting} and EmoFAN \cite{toisoul2021estimation}. As shown in Tab. \ref{tab:multi}, our MTAC performs the best in the discrete emotion classification and obtains competitive CCC scores in the continuous emotion regression on both two benchmarks, which are more robust than another two multi-task methods. One possible reason is that the uncertainty correction of discrete labels optimizes model parameter updates for more discriminative facial features and finally improves generalization performance on the VA estimation task.

\begin{table}[h]
    \centering
    \caption{Multi-task performance comparison. $^*$ denotes using CCC metric$. ^\dagger$ means re-implementing results. \textbf{Bold} denotes the best result, and \textit{ITALICS} indicates the second best result.}\label{tab:multi}
    \setlength{\tabcolsep}{1.5mm}{
    \begin{tabular}{c|c|c|c|c|c}
        \hline
        Dataset & Method & Year & Category & Valence$^*$ & Arousal$^*$ \\
        \hline
        \multirow{6}*{AffectNet} & ResNet-18 & - & 57.94 & 0.672 & 0.608\\
        & DenseNet & - & 59.11 & 0.701 & 0.624\\
        & Emotion-GCN$^\dagger$ \cite{antoniadis2021exploiting} & 2021 & \textit{65.43} & \textbf{0.762} & 0.646 \\
        & EmoFAN$^\dagger$ \cite{toisoul2021estimation} & 2021 & 62.37 & 0.732 & \textbf{0.651} \\
        \cline{2-6}
        & MTAC (ResNet-18) & Ours & 65.09 & 0.753 & 0.635 \\
        & MTAC (DenseNet) & Ours & \textbf{65.80} & \textit{0.758} & \textit{0.649} \\
        \hline
        \multirow{6}*{AffWild2} & ResNet-18 & - & 56.41 & 0.365 & 0.327\\
        & DenseNet & - & 58.70 & 0.406 & 0.352\\
        & Emotion-GCN$^\dagger$ \cite{antoniadis2021exploiting} & 2021 & 62.68 & \textbf{0.451} & \textbf{0.510} \\
        & EmoFAN$^\dagger$ \cite{toisoul2021estimation} & 2021 & 61.70 & 0.429 & 0.496\\
        \cline{2-6}
        & MTAC (ResNet-18) & Ours & \textit{62.78} & 0.446 & 0.487\\
        & MTAC (DenseNet) & Ours & \textbf{63.51} & \textit{0.449} & \textit{0.503}\\
        \hline
    \end{tabular}}
\end{table}

\section{Conclusion}\label{sec:end}
In this paper, we proposed the MTAC framework to alleviate the uncertainty in facial expression images. The target FER branch measured uncertainty to calculate the confidence score and strengthen valid samples during model training. The auxiliary VA branch executed category balancing and joint feature learning with the support of continuous emotion labels. The auxiliary AU branch constructed the data-driven AU graph to generate semantic representations. The relabeling strategy corrected extremely uncertain samples under the feature-level similarity constraint based on the updated memory templates. Our MTAC has a modular design that allows adding and removing branches on the basis what is needed during training and inference. 
Extensive experiments on three large-scale datasets showed that MTAC was found robust to uncertain samples, and achieved superior results in FER task. In the future, other auxiliary tasks such as landmark detection and face recognition can be considered, and MTAC can be extended to generate annotations for unlabeled data, pre-train universal encoders of facial expressions, and address uncertain problem in other data modality.



\section*{Acknowledgment}
This work was supported by the Academy of Finland for the Academy Professor project EmotionAI (grants 336116, 345122), and the Ministry of Education and Culture of Finland for the AI forum project. Dr. Yang Liu also appreciated the support of the China Scholarship Council under Grant 202006150091 in this work. The authors wish to acknowledge CSC-IT Center for Science, Finland, for generous computational resources.

\ifCLASSOPTIONcaptionsoff
  \newpage
\fi



\bibliographystyle{IEEEtran}
\bibliography{IEEEabrv, ref.bib}
\end{document}